\documentclass[conference]{IEEEtran}
\IEEEoverridecommandlockouts
\usepackage{amsmath,amsfonts}
\usepackage{algorithmic}
\usepackage{algorithm}
\usepackage{array}
\usepackage[caption=false,font=normalsize,labelfont=sf,textfont=sf]{subfig}
\usepackage{textcomp}
\usepackage{stfloats}
\usepackage{url}
\usepackage{verbatim}
\usepackage{graphicx}
\usepackage{cite}
\usepackage{comment}
\usepackage{pifont} 
\usepackage{caption}

\usepackage{multirow} 
\usepackage{booktabs}
\usepackage{caption}
\usepackage[table,xcdraw]{xcolor}

\usepackage{booktabs}
\usepackage{siunitx}       
\usepackage{threeparttable}

\usepackage[utf8]{inputenc}
\usepackage{booktabs}
\usepackage{multirow}
\usepackage{graphicx} 

\usepackage{lipsum}
\usepackage{graphicx}
\usepackage{subcaption}
\usepackage{hyperref}

\usepackage{xspace}
\newcommand{\rev}{{\em Revelation}\xspace}
\newcommand{\sol}{{\em IoT Shepherd}\xspace}

\begin{document}

\title{An LLM-Powered AI Agent Framework for Holistic IoT Traffic Interpretation}

\author{
    Daniel Adu Worae\\
    University of Notre Dame\\
    \texttt{dworae@nd.edu}
    \and
    Spyridon Mastorakis\\
    University of Notre Dame\\
    \texttt{mastorakis@nd.edu}
}

\maketitle

\begin{abstract}
Internet of Things (IoT) networks generate diverse and high-volume traffic that reflects both normal activity and potential threats. Deriving meaningful insight from such telemetry requires cross-layer interpretation of behaviors, protocols, and context rather than isolated detection. This work presents an LLM-powered AI agent framework that converts raw packet captures into structured and semantically enriched representations for interactive analysis. The framework integrates feature extraction, transformer-based anomaly detection, packet and flow summarization, threat intelligence enrichment, and retrieval-augmented question answering. An AI agent guided by a large language model performs reasoning over the indexed traffic artifacts, assembling evidence to produce accurate and human-readable interpretations. Experimental evaluation on multiple IoT captures and six open models shows that hybrid retrieval, which combines lexical and semantic search with reranking, substantially improves BLEU, ROUGE, METEOR, and BERTScore results compared with dense-only retrieval. System profiling further indicates low CPU, GPU, and memory overhead, demonstrating that the framework achieves holistic and efficient interpretation of IoT network traffic.

\end{abstract}

\begin{IEEEkeywords}
IoT, RAG, LLMs, device management, anomaly detection
\end{IEEEkeywords}

\footnote{\textbf{Code and data are available at}: \textit{https://github.com/WadElla/Revelation}}

\section{Introduction}
The Internet of Things (IoT) continues to expand at an unprecedented scale, with billions of interconnected devices generating continuous streams of network telemetry \cite{worae2024unified,statista_iot}. These devices span smart home hubs, industrial sensors, medical monitors, and autonomous vehicles—each with distinct communication protocols, lifecycles, and behavioral patterns. While this heterogeneity drives innovation, it also introduces profound challenges for understanding what is happening on the network: whether a device is malfunctioning, whether a configuration was successful, or whether an observed pattern is benign or malicious.

Traditional approaches to IoT traffic analysis are predominantly detection-oriented. They rely on statistical models, signature databases, or deep learning classifiers to label segments of traffic as benign or malicious. While effective in surfacing certain threats, these methods fall short in two fundamental ways. First, they provide little transparency or context, leaving operators with flat labels or unstructured alerts that offer no semantic grounding. Second, they are functionally narrow, being tuned for threat detection rather than a broader operational understanding. Questions such as 
\textit{“Which DNS queries were observed during abnormal device behavior?”}
\textit{“Which public IPs are associated with known vulnerabilities?”}
\textit{“Which MQTT topics are involved in frequent communication bursts?”} remain unanswerable within conventional frameworks. 

In parallel, advances in language models and retrieval-augmented generation demonstrate strong results on structured reasoning tasks \cite{veturi2024rag, borgeaud2022improving, izacard2022few, lewis2020retrieval, khandelwal2019generalization, yasunaga2022retrieval, yao2023react}. However, language models are not natively suited to the raw, irregular structure of network telemetry. They require contextualization, structuring, and often external enrichment to be reliable in high-stakes settings. Placing a language model directly atop logs or packet traces produces brittle behavior and hallucinated explanations, reflecting a mismatch between model assumptions and operational reality.

This paper presents \rev, a workflow for holistic interpretation of IoT network traffic. \rev transforms packet captures into a structured, semantically enriched corpus that links packets, flows, protocol semantics, and device behavior. The corpus comprises packet-layer and flow-level summaries, fine-tuned transformer outputs for anomaly detection, and selective enrichment for public endpoints utilizing threat intelligence. These artifacts are indexed for retrieval so that explanations can be grounded in traceable evidence.
Interactive analysis is guided by an agent that operates over the indexed corpus of traffic-derived artifacts. At question time, it uses tools to locate relevant evidence in the corpus and, when necessary, to consult authoritative IoT sources. Tool use is essential because a pretrained language model cannot parse captures, perform protocol-aware computation, or access up-to-date intelligence on its own. With tool support, the model can integrate results, draft an answer that is tied to concrete evidence, and adapt to the operator’s intent and network context. When a question falls outside the scope of the indexed corpus yet remains in the IoT domain, the agent performs a targeted web lookup and returns a sourced response that is clearly distinguished from capture-derived findings.

\rev pursues a full-spectrum understanding. It summarizes routine and adversarial behavior, connects observations across packets, flows, and protocol semantics, and aggregates device and endpoint context such as MQTT topics, DNS queries, HTTP methods and paths, and Modbus unit identifiers. Explanations are written in clear prose that emphasizes what happened, why it matters, and what action an operator might take, supporting workflows such as triage, change validation, and compliance reporting.

To frame the technical scope and foundational impact of \rev, we pose the following research questions:

\begin{itemize}
    \item \textbf{RQ1}: What is required to move beyond anomaly detection toward full-spectrum interpretation of IoT traffic, covering benign and malicious behavior across diverse protocols and device roles? 
    \item \textbf{RQ2}: To what extent does an agent that processes and retrieves from structured PCAP representations outperform a language model that lacks PCAP-processing tools on traffic interpretation?
    \item \textbf{RQ3}: Which retrieval configuration yields higher answer quality over the same corpus and prompts: dense retrieval alone or a hybrid approach that combines BM25 with dense embeddings, keyword fallback, and cross-encoder reranking?
    \item \textbf{RQ4}: Can \rev provide accurate, context-sensitive answers while remaining efficient and responsive under realistic workloads, as measured by execution time, CPU and memory usage, GPU memory usage, token counts, and response sizes?
\end{itemize}

In addressing these questions, we contribute a novel architecture and set of capabilities that redefine how IoT network telemetry can be interpreted and operationalized:

\begin{itemize}
    
    \item We present \rev, a workflow that converts packet captures into a structured, queryable corpus for holistic interpretation of IoT traffic. The workflow enables a modular processing and representation stack that produces packet-level and flow-level summaries, integrates a fine-tuned transformer for anomaly reporting, and applies selective threat-intelligence enrichment for public endpoints, supporting analysis across benign and adversarial behaviors, protocols, and device roles. 
    \item We develop an agent-guided question answering mechanism that invokes the right tools at question time: a retrieval-and-answer tool over the indexed corpus for capture-grounded queries, and a focused web-lookup tool for IoT questions when local evidence is weak or out of scope. This design yields precise, appropriately sourced answers that reflect operator intent and current context.
    \item We evaluate \rev on a 160-question, artifact-grounded benchmark spanning four PCAPs and six open models under matched prompts and an identical indexed corpus; a hybrid retriever that combines BM25 and dense embeddings with keyword fallback and cross-encoder reranking consistently outperforms dense-only retrieval across BLEU \cite{papineni2002bleu}, ROUGE-1/2/L \cite{lin2004rouge}, METEOR \cite{banerjee2005meteor}, and BERTScore \cite{zhang2019bertscore}.
    \item We profile deployability through an operational study of execution time, CPU utilization, GPU memory, system memory, token counts, and response sizes in both retrieval settings, demonstrating efficiency suitable for local deployment with open models.
    \item We release the full implementation and source code as open source to support transparency and reproducibility: https://github.com/WadElla/Revelation
\end{itemize}

The rest of our paper is organized as follows: Section II describes the system design of \rev. Section III presents experimental results and an analysis of \rev. In Section IV, we present a literature review of \rev. Finally, we conclude in Section V. 


\section{System Design}
\rev is a workflow for holistic interpretation of IoT network traffic. It transforms a packet capture into a structured, indexed collection of traffic-derived artifacts, including protocol logs, packet-layer records, flow summaries, a transformer-based anomaly report, and selective threat-intelligence annotations, then prepares them for retrieval. Interactive analysis is mediated at question time by an agent that uses tools to search this indexed collection and, when a query is IoT-focused but outside the capture, consults authoritative sources. This section outlines the core components and explains how the workflow supports clear, extensible, and operationally relevant analysis.

\begin{figure*}[h]
    \centering  \includegraphics[width=0.8\textwidth]{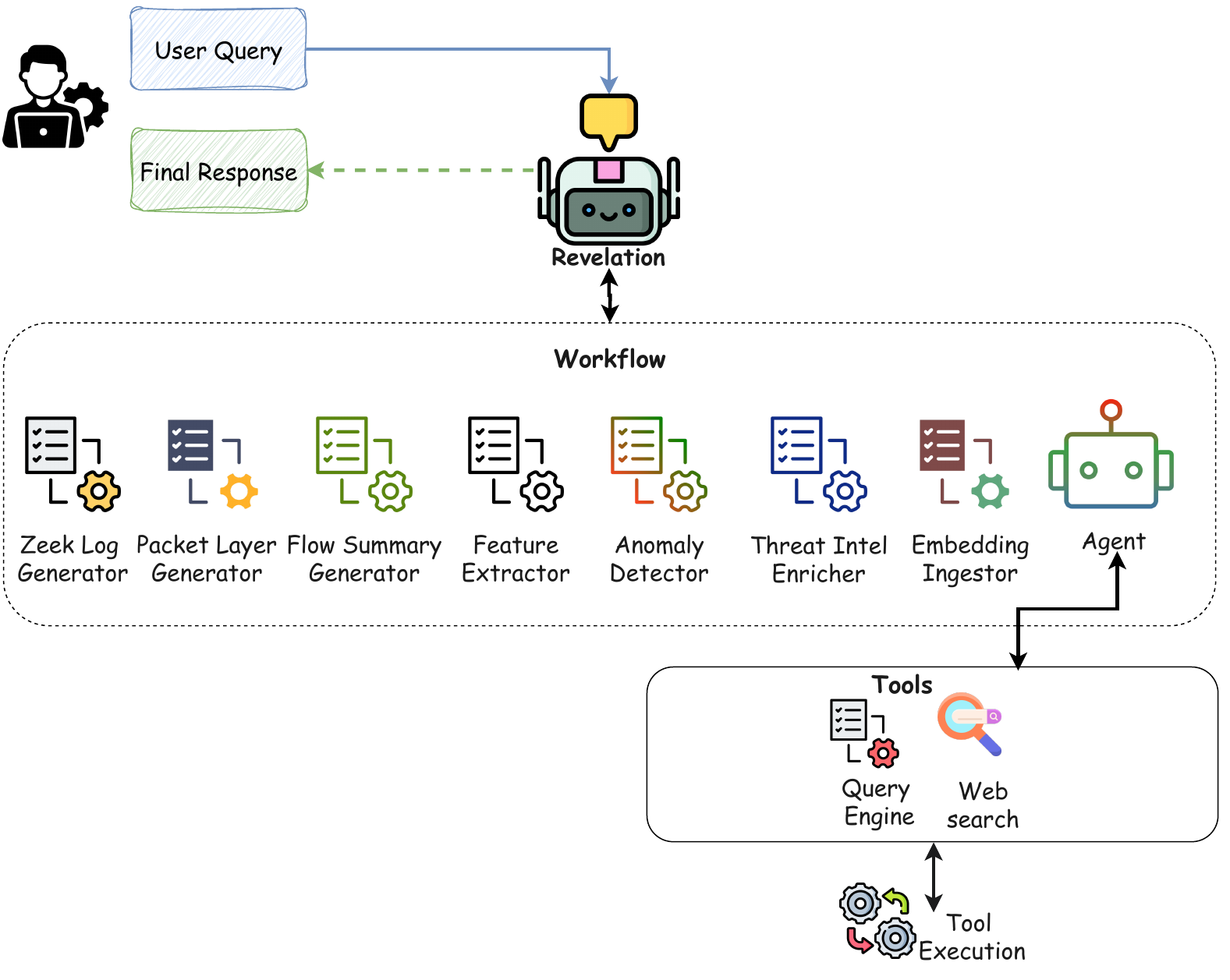} 
    \caption{System Design of \rev}
    \label{fig3}
\end{figure*}

\subsection{PCAP Ingestion and Protocol Log Generation}
\rev begins by applying Zeek to a user-provided PCAP to produce per-protocol JSON logs in Zeek’s standard layout. Zeek is an open-source network analysis framework that reconstructs protocol semantics from packet streams and records them as structured events, widely used in security monitoring and network operations \cite{paxson1999bro}. The logs capture protocol-layer activity common in IoT deployments, including connection events and application transactions for DNS, HTTP, MQTT, Modbus, TLS, and related protocols. Each record preserves original timestamps and salient protocol fields sufficient to characterize requests and responses with fidelity.

The resulting logs are persisted and indexed alongside the transformer-based report, flow summaries, and packet-layer records, forming a searchable representation of protocol behavior for operator-facing question answering. Treating Zeek output as a retrieval surface improves interpretability and supports responses grounded in concrete network events.

\subsection{Packet-Layer Representation}
\rev exports a per-packet JSON view from the PCAP using tshark, capturing the hierarchical structure of headers across the communication stack—link (Ethernet), network (IP), transport (TCP/UDP), and application protocols. Each record preserves the original timestamp, packet order, sizes, and salient protocol fields that characterize an exchange, including TCP control flags and sequence or acknowledgment numbers, and, when present, DNS opcodes, HTTP methods and paths, MQTT control types, and Modbus unit identifiers.

To reduce noise and token overhead, the exporter performs targeted cleaning that removes opaque, high-entropy payload bytes with limited interpretive value, such as raw TCP segment data and TLS randomness, while retaining fields that convey protocol semantics. The result is a normalized packet record that maintains fidelity to protocol structure without extraneous binary content.

The packet-layer JSON is then segmented by semantic chunking to form coherent, retrieval-ready units aligned with conversational boundaries and protocol events. These segments are embedded and indexed alongside protocol logs, flow summaries, and the transformer-based report, enabling fine-grained retrieval and multi-resolution reasoning across the entire communication stack, including handshake reconstruction, timing analysis, and confirmation of specific application exchanges.

\subsection{Flow Summary Generator}
The flow summarizer reconstructs bidirectional conversations from the PCAP across TCP and UDP, treating a flow as the session between two endpoints that share a transport protocol and port pairing. The reconstruction captures both temporal and structural properties of each session so that communication patterns can be analyzed at the level where operators reason about behavior.

For every flow, the summarizer records source and destination addresses, port numbers, transport protocol, total packet count, cumulative byte volume, start and end times, and duration. When application-layer indicators are available, they are incorporated to provide functional context. Examples include HTTP methods and paths, MQTT control messages and topics, DNS queries and responses, Modbus function codes and unit identifiers, and BACnet service invocations.

Connection dynamics are characterized by decoding TCP flag sequences to identify handshake progression, midstream resets, and premature termination. This yields concise signatures of flow health and control behavior that complement application-layer evidence.

Public endpoints observed in a flow are checked against AbuseIPDB \cite{Abuse}, and the summary is annotated inline with the resulting reputation, using abuse-confidence scores to distinguish benign from potentially malicious activity. In addition, MAC addresses are resolved to vendor names via OUI parsing to aid device identification and attribution whenever link-layer information is present.

The output is a human-readable, structured summary in which each flow appears as a standalone narrative block. Summaries are segmented into retrieval-ready units, embedded, and indexed alongside protocol logs, packet-layer records, and the transformer-based report. By representing session-level behavior with application cues, endpoint reputation, and connection-state signals, these summaries enable fine-grained retrieval and support forensic investigation, incident triage, and threat response.

\subsection{Feature Extractor}
This stage derives model-ready features directly from the packet capture using protocol-aware inspection. It computes a fixed set of statistical, temporal, and protocol-specific attributes relevant to IoT environments, drawing from TCP/IP as well as application protocols frequently observed in practice, including DNS, MQTT, and Modbus.

The capture is transformed into a tabular representation in which each row corresponds to a packet or a reconstructed flow, and each column encodes a well-defined feature. Examples include traffic descriptors such as packet and byte counts and durations, transport indicators such as TCP control flags, and application-level fields when present, including DNS query types, MQTT control messages, and Modbus function codes.

For transformer-based inference, each row is then serialized into a single textual sequence of feature–value pairs. Every column receives a descriptive label, followed by a delimiter and the observed value, and pairs are concatenated in a canonical order to form one compact input string. For example, a source address may appear as tcp.dstport:442, with subsequent pairs appended using consistent separators. This textualization preserves the semantics of the tabular representation, enables straightforward handling of mixed data types, and provides a uniform interface for language models to process packet- and flow-level context.

\subsection{Semantic Anomaly Detection and Reporting}
This stage applies a fine-tuned transformer (BERT) to the textual feature sequences produced earlier to classify traffic as benign or as one of several attack types spanning reconnaissance, injection, denial-of-service, and persistence. Beyond per-sample labels, it generates an interpretation report that summarizes traffic composition, identifies dominant threats, and explains observed behaviors in clear prose. Per-attack sections highlight protocol cues and concise operator guidance, and aggregate contextual metadata such as communicating IP pairs, MQTT topics, DNS queries, and Modbus unit identifiers.

\textit{Model training and configuration.} The detector is fine-tuned for IoT anomaly detection using realistic datasets representative of operational conditions, with Edge-IIoTset as the source. Edge-IIoTset covers fifteen attacks across five categories, including TCP SYN flood, port scanning, DNS spoofing, SQL injection, and ransomware, providing breadth across IoT-relevant protocols and behaviors \cite{ferrag2022edge}. Preprocessing removes high-cardinality or dataset-specific identifiers and redundant raw-byte fields to improve generalization. As described in Section D, each data row is serialized into a single sequence of labeled feature–value pairs and tokenized with the BERT tokenizer. Inputs are padded or truncated to a fixed length and passed to a sequence-classification head sized to the number of classes. Fine-tuning follows supervised learning with standard metrics, reporting accuracy, precision, recall, F1, and class-wise performance via a detailed classification report.

\textit{Outputs.} The module emits both a structured CSV of predictions and a narrative report designed for operator use. The report provides class-level summaries with protocol evidence and compiles endpoint and application metadata per detected attack type. These artifacts support situational awareness and serve as structured inputs for subsequent enrichment and retrieval.

\subsection{Threat Intelligence Enrichment}
The enrichment stage identifies public IP addresses observed in the interpretation report and augments them with external context. Two sources are queried: VirusTotal \cite{Virustotal}, which provides aggregate malware-detection counts and recent activity signals, and Shodan InternetDB \cite{Shodan}, which reports exposed services, observed open TCP/UDP ports, descriptive tags, and known CVE identifiers where applicable. Each public IP is mapped to its associated attack category, and a concise, structured summary is inserted into the corresponding metadata subsection of the report.

This contextualization links detected behavior to real-world exposure. For example, an address implicated in scanning may show open SSH and HTTP services, while an address associated with a backdoor class may appear on multiple blocklists with recent malware flags. Enrichment is applied only when public IPs are present, avoiding unnecessary queries while preserving analytical depth.

The enriched report is persisted and indexed with the other traffic-derived representations—protocol logs, packet-layer records, flow summaries, and anomaly outputs—so that retrieval can surface both the behavioral finding and its external intelligence in a single response.

\subsection{Vectorization and Semantic Ingestion}
To support semantically informed reasoning across IoT traffic, the semantic ingestion layer converts protocol logs, anomaly interpretation reports, flow summaries, and packet-layer JSON into a unified vectorized knowledge space. Inputs are segmented with structure-aware chunking and stored as searchable units in a persistent, session-aware vector database. This harmonized representation enables cross-layer querying and traversal of traffic contexts with both breadth and depth.

\subsubsection{Unified Knowledge Integration}
The ingestion pipeline combines four complementary sources of network intelligence, each capturing a distinct dimension of the traffic landscape:
\begin{itemize}
    \item \textbf{Protocol-Level Telemetry:} Structured logs derived from Zeek provide low-level visibility into protocol activity across DNS, HTTP, MQTT, Modbus, and others. These records are chunked using session identifiers to ensure that temporal continuity and logical event boundaries are preserved. This enables downstream reasoning over discrete sessions, message exchanges, and protocol-specific behaviors.
    \item \textbf{Anomaly Interpretation Reports:} Natural language reports generated by the BERT-based anomaly detection module encapsulate anomalous behaviors, threat classifications, and associated metadata, augmented with public threat intelligence from platforms such as VirusTotal and Shodan. These reports are segmented into semantically coherent sections, including summaries, per-attack narratives, and associated metadata, enabling precise alignment between analytic conclusions and their supporting evidence.
    \item \textbf{Narrative Flow Summaries:} Each flow-level summary captures directional communication patterns and temporal structure, enriched with metadata such as protocol usage, port activity, MAC vendor identity, TCP flag sequences, and endpoint reputation derived from external threat intelligence platforms. Flows are segmented individually, while a global summary block contextualizes the overall distribution of traffic across devices and protocols. 
    \item \textbf{Packet-Layer Feature Views:} Extracted from raw PCAP files using deep packet inspection and protocol decoding, this source captures hierarchical protocol fields and structural indicators at the packet level. Semantic chunking is applied using embedding-aware methods to yield context-preserving segments that reflect packet behavior, handshake structures, and multi-layer interactions.
\end{itemize}

\subsubsection{Chunking and Metadata Schema}
To support retrieval-augmented reasoning over heterogeneous IoT traffic representations, our framework applies modality-aware chunking tailored to the structural semantics of each data source:

\begin{itemize}
    \item {\textit{Zeek Logs}:} Logs are segmented using \textit{uid-based} session boundaries. This preserves protocol-level continuity across connection, DNS, HTTP, and other log types, enabling coherent reconstruction of multi-stage behaviors.
    \item {\textit{Anomaly Interpretation Report}:} Generated summaries are chunked by functional sections (e.g., global traffic stats, behavioral narratives per attack type, endpoint metadata enriched with external threat intelligence).  This aligns with the logical flow of the interpretation pipeline and facilitates targeted retrieval during security queries.
    \item {\textit{Flow Summaries}:} Each flow block from the packet capture is treated as an atomic unit, capturing key indicators such as MAC vendors, TTL range, abuse reputation of source and destination IPs, and protocol-level behavior for high-level traffic interpretation.
    \item {\textit{Packet-Layer Structures}:} Raw packets converted to JSON are semantically split using embedding-based chunking (e.g., via \textit{SemanticChunker}). This enables deep inspection of packet-level details without being tied to rigid size-based segmentation.
\end{itemize}

Each chunk is annotated with a compact metadata schema encapsulating the source modality, abstraction level, and a stable internal identifier for deduplication and traceability. The schema preserves contextual integrity across modalities while preventing disclosure of network-identifiable attributes.

\subsubsection{Embedding, Storage, and Session Control}
Following modality-specific segmentation, each chunk is transformed into a dense semantic vector using a locally hosted, domain-optimized embedding model. This model ensures uniform representational quality across heterogeneous inputs, ranging from structured Zeek telemetry and narrative threat reports to flow-level descriptors and packet-layer features, while supporting high-throughput encoding suitable for operational deployments.

The resulting embeddings, paired with essential metadata, are stored in a session-isolated vector database built on Chroma. Each ingestion run generates a uniquely versioned session directory, ensuring reproducibility, provenance tracking, and comparative analysis across temporal states of the network. This session-based organization supports both real-time introspection and historical reconstruction without entangling past and current knowledge representations.

To maintain operational efficiency, the system retains only the three most recent vectorization sessions, automatically purging older instances using a least-recently-used (LRU) eviction policy. This bounded retention strategy enables continuous ingestion and reasoning over dynamic traffic environments while preserving a lightweight storage footprint and minimizing indexing overhead.

\subsubsection{Deduplication and Session Management}
To avoid redundant computation and ensure efficient reuse of knowledge, \rev employs a two-tier strategy that spans both pipeline execution and vector ingestion.

\textit{Pipeline skip guard.} Before executing heavy processing, the framework computes a stable content hash of the input PCAP and compares it with the most recently committed state. If unchanged, upstream processing (e.g., log generation, feature extraction, reporting) is skipped, and previously produced artifacts remain authoritative.

\textit{Ingestion reuse.} Prior to chunking and embedding, the ingestion layer computes file-level content hashes across all input sources before chunking. These hashes are compared against those from the most recent three sessions. Suppose an identical set is detected, indicating that the data has already been indexed. In that case, the ingestion process is skipped, and the latest-session pointer is updated to reference the corresponding session directory. This ensures that even re-uploaded but unchanged traffic artifacts do not incur additional embedding or indexing overhead, while still allowing the query system to route requests to the correct context.

\textit{Session state and reuse.} A lightweight session index maps stable dataset identifiers (e.g., PCAP content hashes) to their associated vector stores and creation times. Interactive queries resolve the active session through this index, ensuring that retrieval operates on the latest consistent context while still allowing explicit selection of historical sessions for side-by-side analysis.


\begin{figure*}[h]
    \centering  \includegraphics[width=0.8\textwidth]{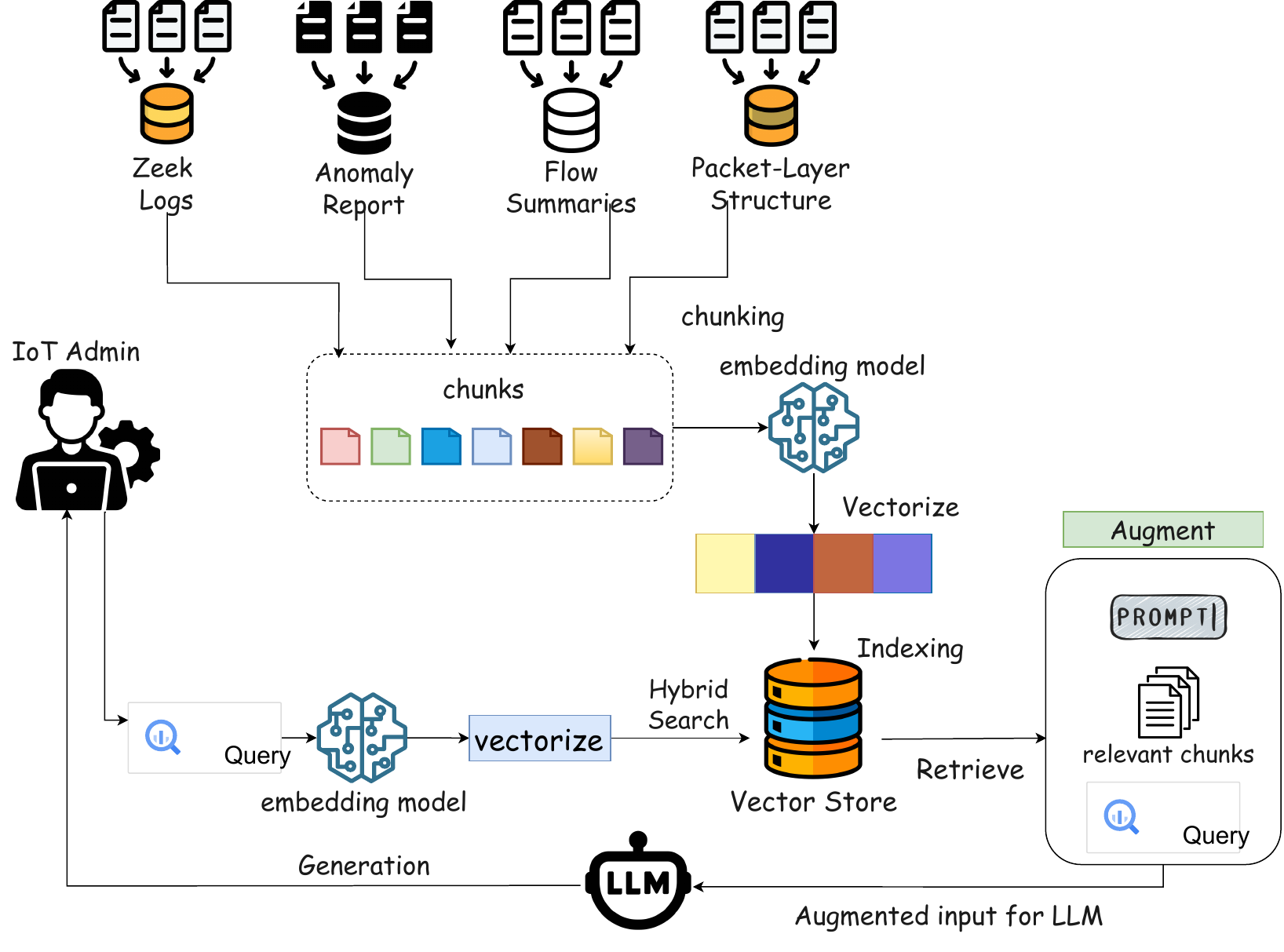} 
    \caption{Retrieval-Augmented Generation Pipeline}
    \label{fig4}
\end{figure*}

\subsection{Contextual Retrieval and Query Engine}
The query engine links natural-language questions to the indexed representations constructed during ingestion, enabling interactive threat forensics, contextual troubleshooting, and operator guidance over heterogeneous IoT traffic. It executes a multi-stage retrieval workflow that combines dense semantic search with BM25 lexical matching and keyword fallback, performs deduplication and cross-encoder reranking, and returns a top-k bundle of evidence drawn from packet-layer records, flow summaries, protocol logs, and interpretation reports (Figures 2 and 3). Retrieval is session-aware so that queries target the most recently processed capture by default. The ranked context is then handed to the agent for answer synthesis.

\begin{figure}[h]
    \centering
    \includegraphics[width=3.45in]{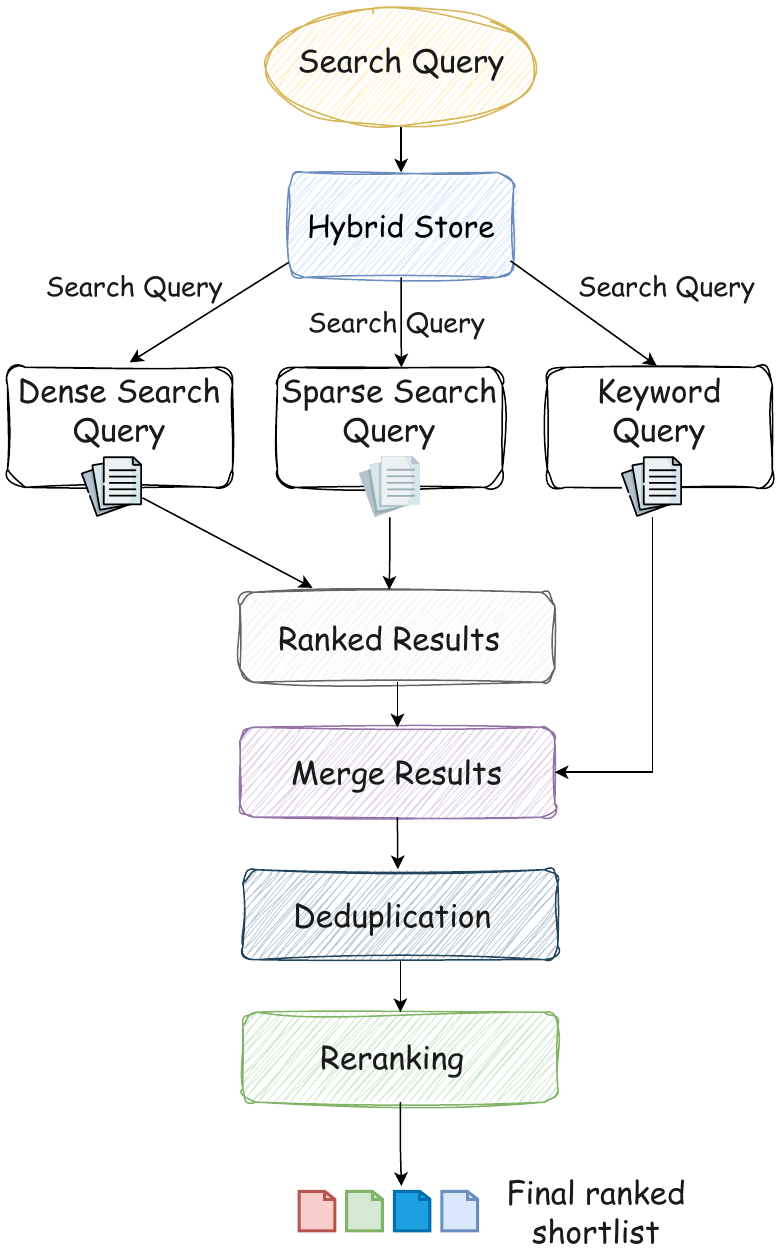} 
    \caption{Hybrid Semantic Retrieval and Reranking Workflow}
    \label{fig3}
\end{figure}

\subsection*{\textbf{Multi-Stage Retrieval Pipeline}}

\subsubsection{Query Vectorization}
User queries are embedded using the same domain-optimized encoder employed during ingestion, producing vector representations that capture both the surface form and latent semantics of the input. This alignment guarantees that abstract queries (e.g., “How is MITM behavior reflected in the traffic metadata?”) can be mapped meaningfully to technical evidence across packet-layer records, flow summaries, protocol logs, and interpretation reports.

\subsubsection{ Hybrid Semantic Retrieval} To maximize retrieval fidelity, the engine performs a hybrid search that integrates three complementary signals, as shown in Figure 3:

\begin{itemize}
    
    \item \textbf{Dense Semantic Search}:  Using cosine similarity, the query vector is compared to stored chunk embeddings. This captures latent conceptual relationships, enabling the system to surface results even when terminology differs (e.g., “unauthorized access” vs. “intrusion”).
    \item \textbf{Sparse Lexical Search (BM25)}: Simultaneously, a BM25 index built over the same document corpus retrieves chunks based on token overlap and term frequency. This enhances precision, particularly for structured logs or keyword-driven forensic queries.
    \item \textbf{Literal Keyword Matching (Fallback)}:  As a robustness mechanism, the engine also applies direct keyword matching on chunk contents. This ensures coverage in cases where both dense and sparse models fail due to edge-case phrasing or rare entity references.
\end{itemize}

Dense and sparse scores are interpolated using a tunable weighting function, and the resulting top-ranked hybrid candidates are merged with fallback matches from the keyword stage to form a comprehensive evidence set. This set is passed downstream for ranking.

\subsubsection{Evidence Consolidation and Reranking} Candidate chunks from the hybrid retrieval stage are deduplicated using content-based hashes to eliminate redundancy. A cross-encoder reranker then performs pairwise relevance scoring between the query and each chunk. This step models deep inter-sentence interactions, producing a final ranked shortlist of the most contextually relevant evidence for the query. 


\subsubsection{Session Awareness and Temporal Consistency} The Query Engine maintains alignment with the most recent ingestion session to ensure that retrieval reflects the latest observed traffic state. Session boundaries are automatically tracked, and historical sessions are rotated to preserve efficiency without sacrificing temporal resolution.

\subsection{Agent for Question Answering}
The agent constitutes the reasoning layer that transforms an operator’s query, together with the ranked context returned by retrieval, into a precise, well-sourced answer. Operating over the active session, it ingests the top-k evidence selected by the query engine and treats this material as its working state. Rather than issuing a one-shot response, the agent executes a compact perceive–reason–act cycle that incrementally resolves the query while remaining aligned with available context.

\textit{Planning is deliberative and revisable.} The agent employs a language model to infer intent, decompose composite requests, and identify residual gaps after initial retrieval. When the evidence set is sufficient, planning reduces to consolidation and articulation of the key points. When coverage is incomplete, the agent specifies narrowly targeted follow-ups—refining retrieval with more discriminative terms or, for IoT-domain details that fall outside the processed capture, consulting an authoritative external source.

\textit{Actions are realized through a bounded toolset.} The agent employs a bounded toolset that aligns with its operational scope. Its default action is to invoke the retrieval-and-answer tool, which merges top-ranked evidence from the indexed corpus and drafts an answer based on the relevant chunks. When the query is IoT-focused but clearly beyond the scope of the capture (for instance, asking about protocol standards, mitigation strategies, or CVE details), or when retrieved evidence proves insufficient, the agent invokes a complementary web-lookup tool. This targeted query ensures that external, authoritative sources are consulted in a controlled manner, yielding citable snippets that complement locally indexed data without diluting precision.

Before finalization, the agent conducts lightweight faithfulness checks. These verify that each claim is supported by the cited passages, ensure consistency of protocol terminology and identifiers, and screen for unwarranted generalization. If support is lacking or ambiguity persists, the agent revises its plan and repeats the minimal necessary retrieval or lookup. The process terminates when the response is adequately supported, appropriately sourced where external knowledge is used, and directly responsive to operator intent.

Agent instructions. The agent is initialized with concise system-level instructions that (i) treat the top-k retrieved context as primary evidence and avoid unsupported speculation; (ii) answer capture-grounded queries using only the provided context, stating explicitly when information is unavailable; (iii) allow pretrained knowledge for clearly general IoT questions when the claim is stable and time-insensitive (e.g., protocol semantics, standard ports); (iv) require a focused web lookup and brief citation for time-varying or source-dependent facts; and (v) enforce plain-text, technically precise, and concise responses, finalized only after a self-check confirms support. At synthesis time, the agent governs orchestration using its internal instructions. When it invokes the retrieval-and-answer tool, it receives a package that has the system prompt, the user query, and the top-k context to frame the Large language model’s behavior.

This design provides disciplined autonomy at query time. By reasoning over session-scoped context, selecting minimal actions to close evidence gaps, and enforcing faithfulness and attribution, the agent produces concise, defensible answers aligned with the semantic representations prepared earlier.

\section{Experimental Results and Analysis}
This section evaluates \rev, an AI agent-powered framework for interpreting IoT traffic. We evaluate along two complementary axes. First, we assess question answering grounded in PCAP-derived artifacts, comparing dense retrieval to a hybrid configuration that combines BM25, dense embeddings, keyword fallback, and cross-encoder reranking. Second, we evaluate semantic anomaly detection using a fine-tuned BERT classifier, reporting accuracy, precision, recall, F1 score, and class-wise summaries. This dual study examines the interpretive reasoning layer and the detection backbone, demonstrating how structured traffic representations and retrieval design impact the quality of answers. At the same time, the classifier provides reliable threat labeling for operational use. 

\subsection{PCAP-Grounded Question Answering}

\begin{table*}[ht]
    \centering
    \captionsetup{justification=centering}
    \caption{Performance Evaluation of \textit{Revelation}'s Retrieval Variants across IoT PCAPs}
    \label{tab:qa_evaluation}
    \resizebox{\textwidth}{!}{%
    \normalsize
    \begin{tabular}{@{}llcccccccccccc@{}}
        \toprule
        \textbf{PCAP} & \textbf{Metric} & \textbf{Gemma2 (Dense)} & \textbf{Gemma2 (Hybrid)} & \textbf{Llama 3.2 (Dense)} & \textbf{Llama 3.2 (Hybrid)} & \textbf{Mistral (Dense)} & \textbf{Mistral (Hybrid)} & \textbf{LLaVA (Dense)} & \textbf{LLaVA (Hybrid)} & \textbf{Qwen3 (Dense)} & \textbf{Qwen3 (Hybrid)} & \textbf{Phi-4-mini (Dense)} & \textbf{Phi-4-mini (Hybrid)} \\
        \midrule
        \multirow{4}{*}{\textbf{Normal traffic}} & BERT (p/r/f) & 79.83 / 84.97 / 82.29 & 86.52 / 85.34 / 85.87 & 80.10 / 85.61 / 82.73 & 86.28 / 88.31 / 87.24 & 81.60 / 85.99 / 83.71 & 87.33 / 88.60 / 87.91 & 84.14 / 87.39 / 85.69 & 88.13 / 89.63 / 88.82 & 77.00 / 84.94 / 80.76 & 78.40 / 85.91 / 81.96 & 69.56 / 74.39 / 71.85 & 84.10 / 87.76 / 85.85 \\
        & ROUGE (r1/r2/rL) & 14.68 / 5.45 / 11.37 & 36.34 / 23.22 / 34.43 & 14.86 / 6.03 / 12.24 & 41.14 / 24.44 / 35.94 & 17.26 / 7.83 / 14.41 & 42.12 / 27.11 / 38.17 & 28.97 / 14.63 / 25.44 & 49.97 / 33.21 / 45.98 & 3.39 / 1.56 / 3.04 & 10.20 / 6.36 / 9.57 & 3.40 / 0.05 / 2.95 & 30.92 / 17.00 / 26.72 \\
        & BLEU & 0.93 & 11.37 & 1.22 & 9.06 & 1.64 & 13.24 & 4.04 & 15.55 & 0.21 & 2.15 & 0.10 & 5.69 \\
        & METEOR & 16.23 & 21.51 & 17.03 & 33.61 & 18.81 & 34.23 & 25.67 & 38.83 & 6.24 & 19.77 & 1.48 & 30.23 \\
        \midrule
        \multirow{4}{*}{\textbf{Uploading attack}} & BERT (p/r/f) & 79.78 / 85.39 / 82.46 & 86.36 / 85.65 / 85.96 & 79.85 / 85.67 / 82.63 & 85.76 / 88.13 / 86.87 & 81.75 / 86.58 / 84.07 & 86.44 / 89.46 / 87.88 & 84.34 / 87.55 / 85.89 & 87.35 / 88.84 / 88.04 & 76.60 / 84.98 / 80.56 & 78.52 / 86.11 / 82.12 & 71.47 / 78.11 / 74.60 & 83.82 / 86.31 / 84.97 \\
        & ROUGE (r1/r2/rL) & 14.58 / 5.74 / 12.19 & 34.35 / 20.08 / 32.46 & 15.87 / 7.56 / 13.22 & 38.62 / 24.34 / 34.86 & 17.97 / 8.04 / 14.48 & 40.19 / 27.05 / 36.84 & 26.84 / 14.65 / 23.77 & 44.93 / 29.60 / 41.29 & 3.42 / 1.57 / 3.08 & 11.00 / 6.78 / 10.07 & 2.93 / 0.26 / 2.57 & 29.22 / 16.90 / 25.15 \\
        & BLEU & 0.82 & 8.68 & 2.44 & 12.28 & 2.06 & 13.68 & 4.58 & 13.93 & 0.20 & 2.35 & 0.14 & 6.64 \\
        & METEOR & 15.45 & 21.48 & 19.24 & 31.27 & 19.30 & 36.40 & 24.84 & 33.88 & 6.20 & 20.63 & 2.18 & 22.05 \\
        \midrule
        \multirow{4}{*}{\textbf{Backdoor attack}} & BERT (p/r/f) & 80.06 / 84.26 / 82.07 & 86.10 / 84.30 / 85.16 & 81.16 / 86.01 / 83.47 & 86.20 / 87.50 / 86.80 & 82.34 / 85.54 / 83.88 & 86.90 / 88.31 / 87.57 & 84.91 / 87.47 / 86.13 & 89.16 / 88.94 / 89.01 & 77.36 / 84.64 / 80.82 & 78.78 / 85.95 / 82.19 & 71.91 / 77.62 / 74.61 & 83.07 / 85.68 / 84.28 \\
        & ROUGE (r1/r2/rL) & 12.99 / 4.54 / 10.50 & 30.95 / 18.08 / 29.34 & 16.62 / 9.13 / 14.43 & 35.25 / 23.19 / 31.57 & 17.83 / 7.37 / 14.62 & 38.25 / 25.32 / 32.96 & 32.39 / 19.00 / 28.85 & 51.62 / 36.48 / 48.19 & 3.91 / 2.10 / 3.67 & 8.88 / 5.61 / 8.11 & 2.13 / 0.14 / 1.83 & 24.07 / 13.98 / 21.27 \\
        & BLEU & 1.06 & 6.28 & 4.66 & 11.60 & 1.87 & 13.46 & 8.40 & 16.35 & 0.45 & 1.73 & 0.09 & 4.66 \\
        & METEOR & 16.58 & 18.71 & 21.86 & 31.95 & 21.69 & 37.65 & 31.53 & 42.97 & 7.88 & 17.52 & 1.46 & 18.43 \\
        \midrule
        \multirow{4}{*}{\textbf{DDoS HTTP flood}} & BERT (p/r/f) & 79.78 / 84.67 / 82.13 & 85.40 / 85.66 / 85.49 & 79.94 / 84.97 / 82.34 & 85.11 / 87.52 / 86.26 & 81.58 / 85.80 / 83.61 & 84.90 / 87.31 / 86.07 & 83.74 / 87.66 / 85.61 & 86.45 / 88.31 / 87.34 & 76.70 / 84.69 / 80.48 & 78.45 / 85.12 / 81.62 & 63.24 / 67.03 / 65.01 & 72.77 / 76.66 / 74.61 \\
        & ROUGE (r1/r2/rL) & 13.26 / 4.75 / 11.10 & 31.27 / 17.68 / 29.57 & 13.67 / 5.61 / 11.25 & 34.00 / 19.24 / 30.10 & 17.53 / 7.77 / 14.66 & 32.04 / 17.45 / 27.81 & 31.72 / 16.49 / 28.60 & 44.92 / 25.72 / 39.98 & 3.78 / 1.75 / 3.51 & 10.38 / 5.59 / 9.25 & 1.75 / 0.00 / 1.32 & 20.37 / 12.58 / 18.51 \\
        & BLEU & 0.80 & 8.18 & 1.18 & 9.54 & 1.92 & 8.10 & 4.17 & 12.60 & 0.27 & 1.75 & 0.08 & 4.69 \\
        & METEOR & 15.10 & 20.47 & 18.40 & 30.13 & 20.76 & 29.85 & 29.70 & 39.32 & 7.48 & 19.41 & 0.34 & 18.50 \\
        \midrule
        \bottomrule
    \end{tabular}}
    \caption*{Performance Metrics: p/r/f = Precision/Recall/F1-Score; ROUGE (r1/r2/rL) = ROUGE-1/ROUGE-2/ROUGE-L.}
\end{table*}

\begin{table*}[ht]
    \centering
    \captionsetup{justification=centering}
    \caption{System Performance and Token Metrics Analysis}
    \label{tab:resource_usage}
    \resizebox{\textwidth}{!}{%
    \begin{tabular}{@{}lcccccccccccc@{}}
        \toprule
        \textbf{Metric} & \textbf{Gemma2 (Dense)} & \textbf{Gemma2 (Hybrid)} & \textbf{Llama 3.2 (Dense)} & \textbf{Llama 3.2 (Hybrid)} & \textbf{Mistral (Dense)} & \textbf{Mistral (Hybrid)} & \textbf{LLaVA (Dense)} & \textbf{LLaVA (Hybrid)} & \textbf{Qwen3 (Dense)} & \textbf{Qwen3 (Hybrid)} & \textbf{Phi-4-mini (Dense)} & \textbf{Phi-4-mini (Hybrid)} \\
        \midrule
        \textbf{Execution Time (s)} & 3.1433 & 5.4415 & 1.5571 & 5.2264 & 1.9995 & 5.6687 & 1.5144 & 6.4263 & 18.4219 & 11.7037 & 3.3381 & 10.0193 \\
        \textbf{Memory Usage (MB)} & 0.0575 & 7.7733 & 0.0497 & 7.7659 & 0.0590 & 7.8483 & 0.0606 & 7.8471 & 0.0544 & 7.8519 & 0.0645 & 7.7780 \\
        \textbf{GPU Memory Used (MB)} & 0.0000 & 16.9607 & 0.0000 & 16.9607 & 0.0000 & 16.9607 & 0.0000 & 16.9607 & 0.0000 & 16.9607 & 0.0000 & 16.9607 \\
        \textbf{CPU Utilization (\%)} & 0.0334 & 2.4839 & 0.0638 & 2.6339 & 0.0512 & 2.3634 & 0.0620 & 2.4466 & 0.0267 & 1.4760 & 0.0535 & 2.2077 \\
        \textbf{Avg. number of tokens} & 225.5867 & 35.4645 & 276.7226 & 76.9648 & 179.3652 & 79.0544 & 117.7630 & 46.7922 & 1631.4593 & 582.0061 & 401.2568 & 547.4785 \\
        \textbf{Avg. Response size (bytes)} & 1185.8805 & 164.2447 & 1271.2372 & 361.9596 & 960.5727 & 401.3950 & 627.4910 & 222.8392 & 7089.2024 & 2593.3212 & 2317.1173 & 3608.0175 \\
        \bottomrule
    \end{tabular}}
\end{table*}

\subsection*{\textbf{1) Experimental Setup}}
We analyze four representative PCAPs covering benign and adversarial scenarios: Normal traffic, Uploading attack, Backdoor attack, and DDoS HTTP flood. Each capture is processed through the workflow into four evidence views: Protocol-Level Telemetry, Anomaly Interpretation Reports, Narrative Flow Summaries, and Packet-Layer Feature Views. For each PCAP, forty ground-truth question–answer pairs (ten per source) are compiled, yielding a benchmark of 160 references.

We evaluate six representative open-source models—Gemma2, LLaVA, Llama-3.2, Mistral, Qwen3, and Phi-4-mini. Two retrieval settings are compared under the same prompts: Dense-only, which ranks candidates based on embedding similarity, and Hybrid, which fuses dense retrieval with BM25 lexical search and keyword fallback before applying cross-encoder reranking. 

\subsection*{\textbf{1) Evaluation Metrics}}
In this question-answering and performance experiment, we employ text quality metrics to evaluate the generated answers and system performance metrics to measure efficiency.

\subsubsection*{\textbf{Text Quality Metrics}}

\begin{itemize}
    \item {\textit{BERT (Precision/Recall/F1-Score)}:} Evaluates semantic fidelity between generated answers and references using contextual embeddings. Precision measures the relevance of generated content, recall assesses coverage of reference meaning, and F1 balances the two.
    \item {\textit{ROUGE (1/2/L)}:} Measures content overlap at unigram, bigram, and sequence levels. ROUGE-1 and ROUGE-2 track whether key entities and phrases are included, while ROUGE-L captures preservation of sentence structure and narrative order.
    \item {\textit{BLEU}:} Focuses on n-gram precision with a brevity penalty, rewarding exact reproduction of technical phrases, protocol strings, and values where precision is critical.
    \item {\textit{METEOR}:} Complements BLEU by balancing precision and recall while allowing for synonymy and stemming. It credits semantically correct paraphrases, ensuring robustness to linguistic variation.
\end{itemize}

\subsubsection*{\textbf{System Performance Metrics}}

\begin{itemize}
    \item {\textit{Execution Time (s)}:} The end-to-end latency measured from the moment the operator's query is received to the moment the final, consolidated answer is delivered. 
    \item {\textit{Memory Usage (MB)}:} Resident memory consumed per query.This represents the maximum amount of physical memory (RAM) consumed by the process. 
    \item {\textit{CPU Utilization (\%)}:} Fraction of logical cores used per query. Measures the computational demand placed on the CPU, indicating processor efficiency. 
    \item {\textit{GPU Memory (MB)}:} The peak device memory allocation on the dedicated GPU, specifically tracking the load introduced by the reranking module and any other device-side operations. 
    \item {\textit{Average Tokens and Response Size (bytes)}:} The mean number of output tokens generated per query is a direct measure of the language model's response verbosity, where reductions indicate tighter, more decisive responses.
\end{itemize}

\subsection*{\textbf{2) Results Across PCAPs}}
As summarized in Table I, Hybrid retrieval consistently and substantially outperforms Dense-only across all four PCAPs and six models. When averaged across all scenarios, BERTScore-F1 improves by 5.3\%, while structure-sensitive metrics show even more dramatic relative gains: ROUGE-L improves by 146.6\%, BLEU by 391.7\%, and METEOR by 83.0\%. These results confirm that Hybrid retrieval produces answers that are semantically faithful, structurally complete, and lexically precise, qualities that are essential for trustworthy interpretation.

Equally significant are the reductions in verbosity. Hybrid cuts the average number of tokens by 51.5\% and reduces response size by 45.6\%, yielding answers that are concise and more decisive. This efficiency has direct operational significance: when a user uploads a PCAP and queries the system, \rev provides answers that are not only more accurate but also more concise, enabling administrators to act quickly without having to wade through redundant or verbose text.

The extent of improvements varies across traffic types, highlighting how \rev adapts to different contexts. For Normal traffic, Hybrid improves BERTScore-F1 by 6.3\%, ROUGE-L by 164.3\%, BLEU by 575.7\%, and METEOR by 96.9\%, while tokens fall by 71.6\%. This shows that \rev can summarize benign activity clearly and efficiently, helping operators confirm normal operations with minimal effort. 

For Uploading attacks, Hybrid achieves BERTScore-F1 gains of 5.2\% and significant relative improvements across ROUGE, BLEU, and METEOR. Token reduction is even sharper at 73.4\%, ensuring that exfiltration cues are highlighted succinctly while retaining full accuracy. 

For Backdoor scenarios, the improvements are more modest, with BERTScore-F1 increasing by 4.9\%, ROUGE-L by 135\%, BLEU by 265.1\%, and METEOR by 66.9\%. Token usage decreases by 24.5\%, ensuring more precise answers that highlight persistence mechanisms with less redundancy, supporting analysts who need to track stealthy behaviors.

For DDoS HTTP floods, Hybrid retrieval improves BERTScore-F1 by 4.6\%, ROUGE-L by 126.3\%, BLEU by 544.6\%, METEOR by 76\%, and reduces the number of tokens by 22.4\%. Although the relative improvements in terms of the number of tokens are smaller than in other cases, the system still provides more precise and less verbose answers. This ensures that operators facing large-scale volumetric attacks can extract essential insights without being overwhelmed by redundant descriptions. 

Taken together, these results demonstrate that \rev's retrieval pipeline is not only technically superior but also operationally feasible. Across benign baselines, stealthy persistence, data exfiltration, and volumetric floods, the system consistently generates answers that are more accurate, concise, and aligned with the holistic interpretation needs. This means that when an administrator uploads a PCAP, \rev provides outputs that are both trustworthy in content and practical in form, ensuring that the system can be deployed as a usable tool for interactive IoT traffic analysis.

\subsection*{\textbf{3) System Performance Analysis}}
System behavior, summarized in Table II, shows that the computational overhead of Hybrid retrieval is modest and predictable. Execution time increases by 52.3\%, reflecting the additional retrieval and reranking steps. Memory usage rises by +7.75 MB, GPU memory by +16.96 MB, and CPU utilization by +2.22\%. These absolute increases remain small in the context of administrative systems, confirming that \rev can scale without resource bottlenecks.

The largest efficiency gains are observed in the generation footprint. Tokens are reduced by 51.5\% and response bytes by 45.6\%, significantly lowering verbosity. This not only offsets retrieval overhead but also improves throughput in multi-query workloads, ensuring that users uploading large PCAPs can obtain timely and actionable answers. In some cases, the shorter outputs even compensate for retrieval overhead, producing neutral or faster end-to-end response times.

Hybrid retrieval significantly enhances the accuracy, conciseness, and interpretability of answers while incurring only minor and bounded resource overhead. This makes Revelation well-suited for interactive, PCAP-driven traffic interpretation in real operational settings.

\subsection{Anomaly Detection}
To assess \rev's anomaly detection stage, we fine-tuned a BERT classifier on the Edge-IIoTset dataset, which encompasses fifteen distinct traffic classes covering reconnaissance, injection, denial-of-service, persistence, and benign activity. To ensure balanced representation of all categories, the dataset was partitioned into training and evaluation sets using an 80/20 stratified split, yielding 126,240 samples for training and 31,560 for testing.

\subsection*{\textbf{1) Experimental results}}
The model achieves 99.88\% accuracy, with weighted precision, recall, and F1-score of 99.88\%. This reflects consistent performance across both benign and adversarial traffic, demonstrating strong generalization to unseen flows.

\subsubsection*{\textbf{a) ROC-AUC Analysis}}
Figure 4 presents the per-class ROC curves. Every class attains an area under the curve of at least 0.99; in particular, MITM, DDoS over UDP, and DDoS over ICMP achieve exactly 1.00. Normal traffic records approximately 0.9887, and all remaining attack classes fall between 0.9988 and 1.0000. Collectively, these results indicate near-saturated separability across classes and stable discrimination even in the low false-positive operating region.

\subsubsection*{\textbf{b) Confusion Matrix}}
Figure 5 shows consistently high recalls across classes. Perfect recall (100.00\%) is achieved for Man-in-the-Middle (MITM), Ransomware, SQL Injection, Distributed Denial-of-Service (DDoS) over UDP, and DDoS over ICMP. Near-perfect results are observed for Uploading, DDoS over HTTP, DDoS over TCP, and Port Scanning (greater than or equal to 99.95\% recall). Normal traffic is correctly identified with 99.61\% recall, while the lowest recall, 99.50\%, occurs for Fingerprinting, reflecting minor confusion among closely related reconnaissance flows. Precision remains uniformly high; the lowest class precision is 99.18\% for Ransomware, where recall remains perfect, yielding an F1-score near 99.59\%.

To provide a comprehensive view of class-level performance, Table III reports the full classification metrics, including precision, recall, and F1 Scores across all fifteen traffic categories. The combination of Figures 4 and 5 with Table III demonstrates that the transformer-based anomaly detector achieves highly reliable classification across both benign and adversarial traffic, reinforcing its role as a dependable foundation within the \rev framework.
\begin{figure}[h]
    \centering
    \includegraphics[width=3.52in]{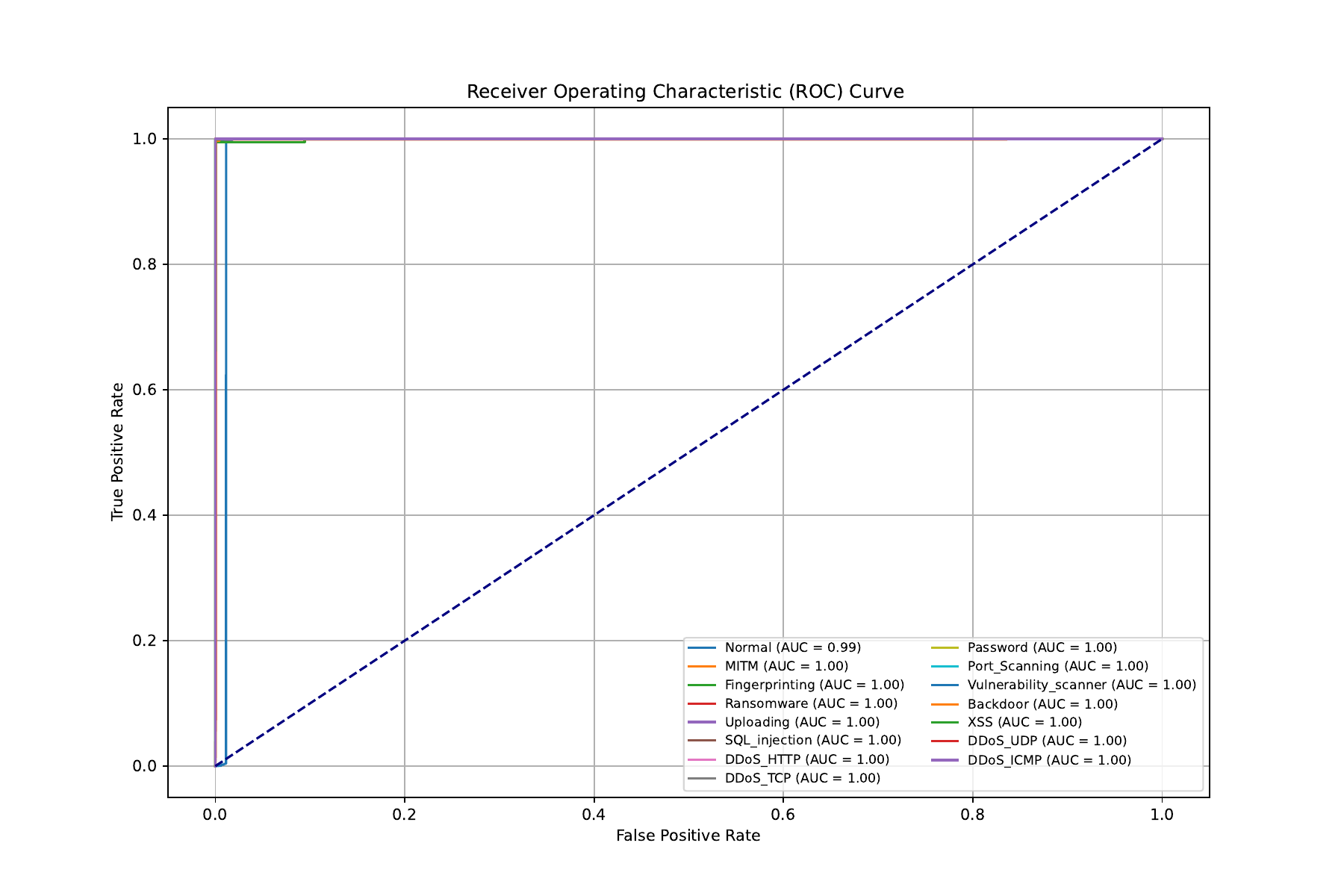}  
    \caption{ROC AUC scores showcasing our fine-tuned BERT model's performance in IoT anomaly detection.}
    \label{fig5}
\end{figure}

\begin{figure}[h]
    \centering
    \includegraphics[width=3.54in]{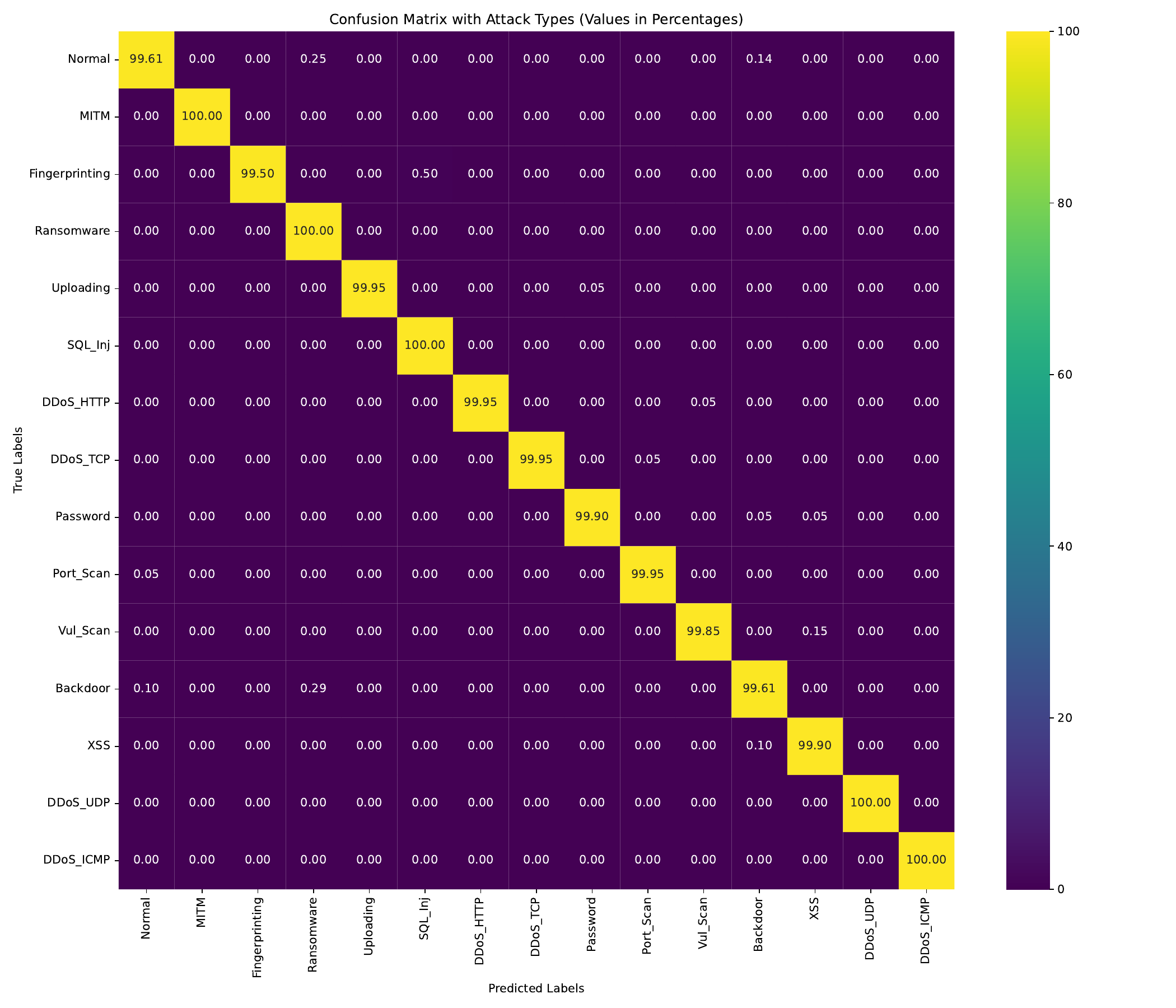} 
    \caption{Confusion matrix showcasing the classification performance of our fine-tuned BERT model for IoT anomaly detection.}
    \label{fig6}
\end{figure}

\begin{table}[t]
    \centering
    \caption{Classification report of the anomaly detector}
    \label{tab:classification_report_revelation}
    \resizebox{\columnwidth}{!}{%
    \begin{tabular}{@{}lcccc@{}}
        \toprule
        \textbf{Class} & \textbf{Precision} & \textbf{Recall} & \textbf{F1-Score} & \textbf{Support} \\
        \midrule
        Normal                & 99.94 & 99.61 & 99.77 & 4,860 \\
        MITM                  & 100.00 & 100.00 & 100.00 & 243 \\
        Fingerprinting        & 100.00 & 99.50 & 99.75 & 200 \\
        Ransomware            & 99.18 & 100.00 & 99.59 & 2,185 \\
        Uploading             & 100.00 & 99.95 & 99.98 & 2,054 \\
        SQL\_Injection        & 99.95 & 100.00 & 99.98 & 2,062 \\
        DDoS\_HTTP            & 100.00 & 99.95 & 99.98 & 2,112 \\
        DDoS\_TCP             & 100.00 & 99.95 & 99.98 & 2,050 \\
        Password              & 99.95 & 99.90 & 99.92 & 1,998 \\
        Port\_Scanning        & 99.95 & 99.95 & 99.95 & 2,014 \\
        Vul\_Scanner          & 99.95 & 99.85 & 99.90 & 2,015 \\
        Backdoor              & 99.51 & 99.61 & 99.56 & 2,039 \\
        XSS                   & 99.80 & 99.90 & 99.85 & 2,010 \\
        DDoS\_UDP             & 100.00 & 100.00 & 100.00 & 2,900 \\
        DDoS\_ICMP            & 100.00 & 100.00 & 100.00 & 2,818 \\
        \midrule
        \textbf{Macro Avg}    & 99.88 & 99.88 & 99.88 & 31,560 \\
        \textbf{Weighted Avg} & 99.88 & 99.88 & 99.88 & 31,560 \\
        \textbf{Accuracy}     & \multicolumn{4}{c}{\textbf{99.88\%}} \\
        \bottomrule
    \end{tabular}}
\end{table}

\subsection{Security and Privacy Analysis}

\subsection*{\textbf{Data Protection and Confidentiality}}
\rev safeguards privacy by processing and storing all traffic artifacts locally, including packet captures, logs, flow summaries, and vector embeddings. Only enrichment queries to trusted intelligence sources, such as VirusTotal and Shodan InternetDB, are issued externally. These requests are restricted to confirmed public IP addresses to prevent the disclosure of internal network details. The framework’s reliance on locally hosted open models ensures that sensitive telemetry remains within the administrative perimeter, maintaining full control over data and inference processes.


\subsection*{\textbf{Security and model safeguards}}
The framework enforces strict operational boundaries on both language models and agents to prevent unintended actions or information exposure. The agent’s autonomy is limited to reasoning, retrieval, and controlled tool invocation. It cannot modify configurations, alter network states, or transmit sensitive data. Each operation is logged for accountability, and predefined guardrails ensure that prompts and outputs adhere to factual, context-bound, and safety-conscious standards. These mechanisms collectively uphold system integrity, ensuring that intelligent reasoning supports, rather than endangers, secure network operations.

Together, these measures make \rev a secure and privacy-preserving framework for interpretable IoT traffic analysis.

\section{Related Work}
\label{sec: Lit}
Large language models (LLMs) are beginning to reshape how network data is analyzed and explained. Early studies have tested whether general-purpose LLMs can solve networking tasks such as reasoning about topologies, answering diagnostic queries, or interpreting configurations. These works show encouraging results in small or medium networks, but performance degrades as complexity grows, suggesting that purely language-based reasoning is insufficient without structured context \cite{donadel2024can}. Related efforts treat network traces as text-like inputs, enabling models to narrate packet activity in natural language. While this demonstrates feasibility, the resulting explanations often lack precision because packet data carries limited semantic cues when presented in isolation \cite{de2024exploring}.

To address this gap, researchers have proposed adapting LLMs more directly to traffic data. Approaches such as TrafficLLM introduce traffic-specific tokenization and multi-stage tuning, achieving strong performance in both classification and generation across diverse scenarios, including unseen traffic types \cite{cui2025trafficllm}. Agent-based systems extend this idea by using LLMs to coordinate detection pipelines and generate explanations of intrusion events, reporting competitive accuracy even in zero-day settings \cite{li2024ids}. Other explainable intrusion detection frameworks employ transformer models to attach rationales to anomaly alerts, improving transparency but raising concerns about cost and efficiency \cite{houssel2024towards}. These efforts advance detection, yet they often reduce network activity to a binary or categorical label and provide only limited insight into benign behaviors that dominate real-world traffic.

Parallel work has explored operational analysis of logs and packet captures. Systems have been developed to automate PCAP failure detection under scarce labels \cite{rahman2024leveraging} and to assist security analysts with conversational log summarization \cite{balasubramanian2024cygent}. More comprehensive designs preprocess and enrich PCAPs, embed them in vector databases, and query them with local LLMs for post-hoc incident reconstruction \cite{rahman2024leveraging}. Such approaches show that LLM-guided workflows can support forensic tasks, but they typically operate retrospectively and do not unify anomaly detection with a broader interpretation of ongoing network activity. Prior work in IoT administration has also shown that retrieval-augmented generation improves operator answers when paired with anomaly signals, though the focus was on manuals and device documentation rather than raw traffic itself \cite{worae2024unified}.

Beyond traffic and logs, LLMs have demonstrated effectiveness in security intelligence and domain-specific adaptation. In cyber threat intelligence, schema-guided prompting enables accurate extraction of structured events from noisy forum text \cite{clairoux2024use}. At the vulnerability level, models have been tested on generating CVSS scores for CVEs, where they approximate expert scoring but suffer from consistency issues \cite{marchiori2025can}. In protocol analysis, combining retrieval with step-by-step reasoning improves state-machine inference by generating more effective packet seeds for fuzzing \cite{maklad2025retrieval}. Meanwhile, in telecommunications, both fine-tuned and zero-shot evaluations show that LLMs can capture technical language and intents, although performance is highly variable without specialized training \cite{bariah2023understanding, ahmed2024linguistic}. These advances confirm that domain adaptation and structured inputs can stabilize LLM outputs; however, their scope is limited to textual corpora or specific protocol logic, rather than the complex mixture of logs, flows, and enriched metadata that characterizes real network environments.

The literature shows that LLMs can reason about networks, adapt to traffic-like data, detect anomalies, and extract structured knowledge. What remains absent is a unified framework that interprets IoT traffic end-to-end, explaining both benign and malicious activity, grounding outputs in heterogeneous telemetry and threat intelligence, and producing explanations that operators can act on. \rev addresses this gap.

\section{Conclusion} 
This work presented \rev, an AI-agent–powered framework that transforms raw IoT packet captures into a structured, semantically enriched corpus and enables evidence-grounded question answering. By unifying protocol logs, packet-layer views, flow summaries, a fine-tuned transformer for anomaly detection, and selective threat-intelligence enrichment, \rev supports multi-resolution reasoning that connects routine behavior with malicious activity. An agent mediates interactive analysis by invoking retrieval and web-lookup tools, producing concise and defensible answers to operator queries. Evaluation across four representative PCAPs and six open models shows that hybrid retrieval substantially improves accuracy, structure, and concision compared with dense-only baselines, while resource profiling confirms efficiency suitable for local deployment. \rev thus advances IoT traffic analysis beyond anomaly detection toward interpretable, context-aware, and operationally relevant intelligence.

\bibliography{Revelation}

\bibliographystyle{IEEEtran}

\end{document}